\documentclass[11pt]{article}

\usepackage[preprint]{acl}

\usepackage{times}
\usepackage{latexsym}

\usepackage[T1]{fontenc}

\usepackage[utf8]{inputenc}
\usepackage{enumitem}

\usepackage{microtype}
\usepackage{booktabs}
\usepackage{amsmath}
\usepackage{cleveref}
\usepackage{graphicx}
\usepackage[most]{tcolorbox}
\usepackage{enumitem}
\usepackage{inconsolata}

\title{Don't Stop Early: Scalable Enterprise Deep Research with Controlled Information Flow and Evidence-Aware Termination}

\author{\quad Prafulla Kumar Choubey$^{*}$ \quad Kung-Hsiang Huang$^{*}$  \quad Pranav Narayanan Venkit$^{*}$  \\
 {\bf Jiaxin Zhang$^{*}$ \quad Vaibhav Vats \quad Yu Li \quad Xiangyu Peng  \quad Chien-Sheng Wu}\\
Salesforce AI Research \\
\small{
   \textbf{Correspondence:} \href{mailto:pchoubey@salesforce.com,wu.jason@salesforce.com}{[pchoubey, wu.jason]@salesforce.com}}\\
 \small{
   $^{*}$Authors contributed equally
}\\
}

\begin{document}
\maketitle
\begin{abstract}

Enterprise deep research often fails to produce decision-ready reports due to uneven information coverage, context explosion, and premature stopping. We propose a \textit{scalable} Enterprise Deep Research (EDR) architecture to address these failures. Our system (i) decomposes requests into coverage-driven objectives via outline generation with reflection, (ii) localizes context with dependency-guided execution and explicit information sharing, and (iii) enforces evidence-based completion criteria so agents iteratively collect information until sufficiency conditions are met. We evaluate on an internal sales enablement task and the public DeepResearch Bench benchmark, where our proposed system design achieves the strongest overall performance compared with competitive deep-research baselines. The results show that dependency-controlled context and explicit evidence sufficiency criteria reduce premature stopping and improve the consistency and depth of enterprise research outputs. 

\end{abstract}

\section{Introduction}
Deep Research (DR) systems~\citep{openai2025deepresearch, google2025deepresearch, anthropic_meet_claude_2025} typically generate long, structured reports designed to provide comprehensive coverage of a topic. In enterprise settings, the requirements are significantly more demanding \cite{huang2025crmarena}. Reports are expected not only to be comprehensive but also to deliver deeper, decision-ready analysis that integrates publicly available information with internal organizational knowledge such as customer relationship management (CRM) records, prior deal history, and product documentation \cite{choubey2025benchmarking}.
Consequently, Enterprise Deep Research (EDR) systems must first identify the appropriate source for each required piece of information (internal systems or external data) and then perform multi-step reasoning processes to integrate this information into a coherent, structured analysis.

Beyond information integration, enterprise DR systems face two practical challenges. First, execution efficiency becomes critical. DR tasks naturally involve multiple stages of investigation, and purely sequential execution leads to very long runtimes. In enterprise environments, where timely insights are often required, efficient execution is essential~\citep{lin2026wdscalingparalleltoolcalling}. EDR systems must therefore enable parallel execution of independent research steps while preserving logical dependencies among tasks, ensuring that improvements in execution speed do not compromise coverage or correctness.

Second, enterprise research tasks are evidentially uneven~\citep{zhang2025fargenuinelyusefuldeep}: the availability and quality of supporting information vary significantly across customers, products, and markets. For some areas, relevant data may be abundant; for others, evidence may be sparse, fragmented, contradictory, or difficult to verify. As a result, it is often unclear when sufficient information has been collected to produce decision-ready analysis. In practice, agents often stop after collecting limited surface-level evidence, leading to uneven depth and inconsistent analytical rigor across sections. Robust EDR systems must therefore reason explicitly about evidence sufficiency, identify gaps, and continue targeted investigation until each part of the analysis meets a consistent standard of support~\footnote{Enterprise data access is governed by permission hierarchies, data residency, and compliance requirements. All our research and evaluation were conducted in controlled internal environments; authentication and permission enforcement were therefore not implemented, though privacy and access control remain important for real-world deployment.}.

To address these challenges, we design our DR system around three core principles: 
a) high coverage-driven task decomposition, 
b) easy-to-control context and dependency-based information flow, and 
c) robust evidence-based termination criteria. 
We first convert a research request into a structured outline that enumerates explicit information objectives, ensuring that all analytical dimensions are represented upfront.
We then transform this outline into executable research steps and specify dependencies among them, allowing independent steps to execute in parallel while preserving logical consistency across the overall research process. Each step is executed by a specialized agent operating within a bounded local context, receiving only the inputs necessary for its objective. Explicit dependency links control when intermediate results are shared, preventing uncontrolled context growth. Finally, each step defines sufficiency conditions prior to execution, requiring agents to iteratively gather and evaluate evidence until predefined completion criteria are met, thus reducing premature termination and stabilizing analytical depth across the report.

We evaluate our design on (i) internal enterprise sales enablement tasks involving structured win-card generation and (ii) the public \textit{DeepResearch Bench} benchmark~\citep{du2025deepresearch}. Under matched public-search constraints, our system improves win-card coverage relative to strong DR baselines; enabling internal tools and knowledge yields further gains. On DeepResearch Bench, our approach achieves the best overall average score among compared methods, with especially strong improvements in comprehensiveness and insightfulness. Ablation studies indicate that dependency-gated context sharing and step-level sufficiency criteria are key contributors to performance.

Overall, our results show that \emph{coverage-first decomposition}, \emph{dependency-controlled information flow}, and \emph{evidence-based termination} form practical foundations for enterprise deep research: they reduce omissions, prevent context growth, and limit premature stopping by enforcing consistent standards of evidential support across sections.



\begin{figure*}[t]
    \centering
    \includegraphics[width=\linewidth]{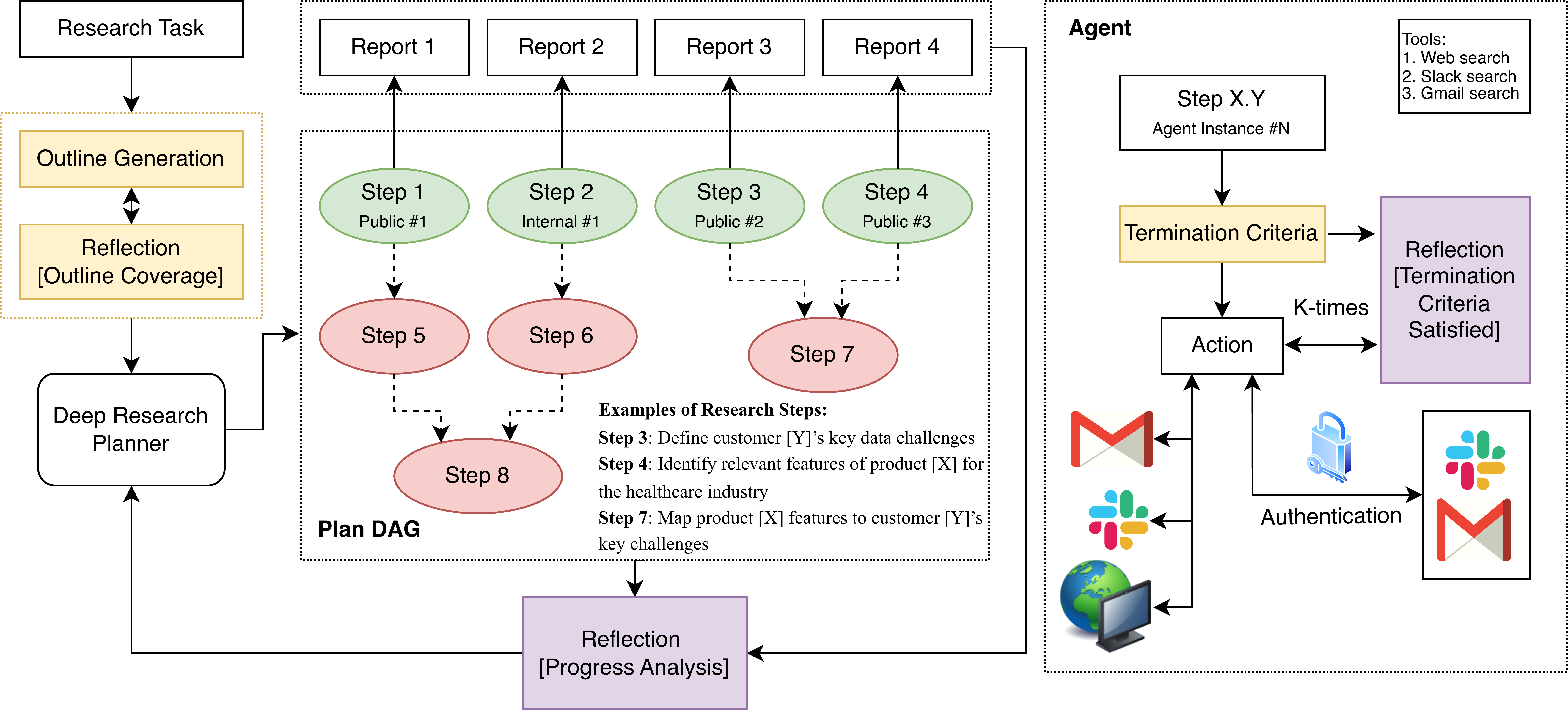}
\caption{Overview of the proposed Enterprise Deep Research (EDR) system.}
    \label{fig:overview}
\end{figure*}

\section{Related Work}
DR systems typically follow one of two designs. Single-agent approaches~\cite{yao2023reactsynergizingreasoningacting, MoonshotAI2025KimiResearcher, tao2025webshaperagenticallydatasynthesizing, nguyen2025sfrdeepresearcheffectivereinforcementlearning, li2025websailornavigatingsuperhumanreasoning} use one model to plan, search, and write in a shared context. This can improve coherence, but it is often costly, hard to steer over long tasks, and prone to context growth as tool outputs and intermediate drafts accumulate. Multi-agent approaches~\citep{alzubi2025opendeepsearchdemocratizing, MiroFlow2025, li2025webthinkerempoweringlargereasoning, LangChainAI_OpenDeepResearch_2025, ByteDance_DeerFlow, openmanus2025, prabhakar2025enterprisedeepresearchsteerable} split work across roles (e.g., planner, searcher, writer, reviewer), which can improve modularity and parallelism. In practice, they often introduce coordination overhead, uneven coverage across sections, and excessive cross-agent communication that inflates context and can lead to early stopping. We build on a multi-agent architecture, but focus our design on reducing these failure modes.

Graph-based planning has also been explored for a range of LLM reasoning and agentic tasks \cite{liu2025graphaugmentedlargelanguagemodel, wang2025agentdropoutdynamicagentelimination, wu2024agentkitstructuredllmreasoning, zhang2025planovergraphparallelablellmagent}, including code generation \cite{wei2025vflowdiscoveringoptimalagentic}, tool-use workflows \cite{wei2025reactplannercentricframeworkcomplex}, and RAG \cite{verma2025planragefficienttesttimeplanning}. Inspired by these efforts, we use a plan DAG to support task decomposition and scheduling, while also controlling context by limiting what information can flow between dependent steps.


Recent analyses of multi-agent LLM systems emphasize that termination is a recurring failure point. \citet{cemri2025multi} identify ``task verification and termination'' as a major error category, where agents stop early or fail to recognize unmet requirements. Most DR systems prioritize improved planning and retrieval, but do not explicitly control termination. In contrast, our approach encodes completion criteria at both planning and agent execution stages, substantially reducing premature stopping in our experiments.

\section{Deep Research System}
We observe three recurring failure modes in naive multi-agent deep research systems~\citep{cemri2025multi}: 
\textit{uneven coverage}, 
\textit{context explosion}, and 
\textit{premature stopping}. 
These issues lead to inconsistent analytical depth and unstable report quality.
We address these problems through three corresponding design mechanisms:

\begin{itemize}[leftmargin=*]

\item \textbf{High coverage-driven task decomposition (uneven coverage):}
Before execution begins, the system explicitly enumerates the objectives from the research task. The plan is constructed to satisfy these objectives rather than being driven by available tools, ensuring that all required objectives are represented upfront.

\item \textbf{Context localization using dependency-based information flow (context explosion):}
Each research step is executed independently with bounded context, accessing only the outputs of the steps it depends on rather than the full execution history. Outputs are shared only when required by explicit dependencies, preventing accumulation of irrelevant context and reducing cascading errors.

\item \textbf{Evidence-based termination criteria (premature stopping):}
Each research step defines explicit sufficiency conditions before execution. Agents gather and evaluate evidence until these conditions are met, reducing early termination and improving consistency in analytical depth.

\end{itemize}

\subsection{System Architecture}

Given a research task, the system converts it into an explicit execution plan and coordinates multiple agent workflows. The architecture 
operationalizes the three design mechanisms described above by separating 
(1) outlining research objectives to ensure coverage, (2) planning investigation steps with dependency-aware execution to localize context, and (3) executing agents with explicit termination criteria to prevent premature stopping (\cref{fig:overview}).


\paragraph{Outline Generation and Reflection:}
The system first generates a structured outline aligned with the research task, expanding each major component into concrete sub-questions that specify the required information. For example, a customer background component may include the company’s industry, business model, recent initiatives, and organizational structure. This process makes the full scope of required information explicit before investigation begins. We then perform an outline reflection step that checks for missing fields, underspecified questions, or ambiguous objectives. Identified gaps are resolved before planning, reducing the likelihood of incomplete coverage during execution.

Our outline generation and reflection are performed without querying any information gathering tools. For research tasks where information is not readily available, early retrieval can bias the process toward tangential but easily accessible details, leading to drift from the core objective. By first constructing a high-level, coverage-driven outline grounded in the task definition, the system establishes a stable frame for subsequent investigation. This design does not restrict retrieval. Instead, it separates task framing from evidence collection, ensuring that downstream planning and execution remain guided by the research objectives rather than the distribution of available evidence.

\paragraph{Planning, Dependency Construction, and Replanning:}
The refined outline is passed to a planner that converts related sub-questions into executable research steps. Each step corresponds to a specific information objective, allowing closely related questions to be addressed within a single investigation.

For each step, the planner assigns an appropriate agent type (public researcher or internal agent), defines dependencies based on information requirements, and determines which steps can execute in parallel. The resulting plan is represented as a directed acyclic graph (DAG); for example, in \cref{fig:overview}, independent step 3 (defining customer [Y]’s key data challenges) and step 4 (identifying relevant features of product [X] for the healthcare industry) can run in parallel, while step 7 (mapping product [X] features to customer [Y]’s key challenges) can only be executed after steps 3 and 4 are completed, since it requires their outputs as inputs.

After completing the currently executable set of steps, the system performs progress reflection by comparing collected outputs against the original outline to identify missing coverage or insufficient evidence. Based on this evaluation, the planner may introduce new steps, refine existing ones, or update dependencies to incorporate newly available information. Planning and replanning are performed for a fixed number of iterations, treated as a system hyperparameter. After the final planning iteration, the plan is frozen, and all remaining steps in the DAG are executed to completion according to their dependency constraints.

\paragraph{Agent Execution and Termination:}
When a step becomes executable, the system calls an agent configured for the corresponding objective. Each agent has access to its own tools and to selected outputs from prior steps, as specified by the planner. This controlled information sharing prevents unrelated steps from inheriting irrelevant context.

Agents execute independently and may run concurrently when dependencies allow. Each agent follows an iterative reasoning-and-action loop. At initialization, the agent defines explicit \textit{termination criteria} that specify what information must be collected to complete the assigned objective. For example, identifying the customer’s organizational structure may require determining relevant teams, key roles, their responsibilities, reporting relationships, and decision ownership.

During execution, the agent uses tools to gather evidence and evaluates whether termination criteria are satisfied. The loop continues until sufficient evidence is collected. Upon termination, the agent produces a structured step-level output that becomes available to dependent steps. This ensures bounded execution while improving consistency in analytical depth across the research task.

\paragraph{Agent Types and Tool Interfaces:}
We define agent types based on the primary information source.
The public researcher agent gathers publicly available information such as company announcements, industry trends, and competitor activity using web-based tools.
The internal agent retrieves internal enterprise data, including CRM records, prior interactions, opportunity context, and internal documentation relevant to the target account.

The planner assigns agent types based on the objective of each step, while tool selection occurs during execution.
Public information is accessed through \textit{Web Search}.
Internal data access is implemented via MCP-based tool interfaces~\citep{anthropic2024mcp}, which expose enterprise connectors for \textit{Enterprise Knowledge Search} over unstructured repositories (e.g., Confluence, Highspot, Trailhead), \textit{CRM Search} for structured customer and opportunity data, and \textit{Conversation Search} for indexed internal communications (e.g., Slack).

We show examples of the outline (\Cref{appendix-outline}), the PlanDAG (\Cref{appendix-plan-dag}), and the agent termination criteria (\Cref{appendix-termination-criteria}).

\section{Enterprise Deep Research Task for Sales Enablement}
Our DR task focuses on automatically generating structured win-cards to support sales preparation. Given a target account such as CVS Health and a predefined multi-section template, the system must produce a complete, account-specific report for customer meetings. The template includes sections such as customer background, customer engagement footprint, organizational structure, business goals and pain points, tailored value propositions, competitive landscape, technical readiness, relevant success stories, and key discovery questions. These sections define broad objectives rather than fixed fields, and the required content varies depending on the product, industry, and customer context.

Completing each section requires investigation and synthesis rather than simple field population, and the reasoning process must adapt to each account. For example, identifying business goals for a large public enterprise may require analyzing earnings calls and regulatory filings, whereas for a private growth-stage company it may rely on hiring trends and product announcements. The system cannot apply the same reasoning steps uniformly across customers. Each section therefore represents a standalone research task that demands context-specific analysis while still conforming to the overall template structure.

\section{Experiments and Results}

\subsection{Experimental Setup}
We evaluate our proposed DR system on 10 randomly selected customer scenarios. Each scenario corresponds to a different target account, while following the same predefined win-card template. This setup allows us to assess whether the system can effectively adapt content to variations in customer context, industry characteristics, and the availability of supporting evidence. We evaluate under two tool configurations: \textit{Public Search Tool}, where only external web search is available; and \textit{Public + Internal Search Tools}, where the system additionally has access to enterprise knowledge search, CRM search, and conversation search tools. Public search is implemented using Tavily Search\footnote{\url{https://tavily.com}}, and the overall system is implemented using LangGraph\footnote{\url{https://github.com/langchain-ai/langgraph}} to support structured multi-agent orchestration and controlled execution flow.

We compare our approach against several strong baselines on both the sales enablement task and DeepResearch Bench. We first include two LLM-based baselines, GPT-4.1 and GPT-5.1, each paired with a web search tool. For these baselines, we evaluate the standalone performance of the underlying models, without additional multi-agent orchestration or context management. For GPT-5.1, we enable heavy-reasoning mode and allow retries until a valid output is produced.

We also compare our approach with both proprietary and open-source Deep Research (DR) baselines on the sales enablement task. The proprietary baselines include Gemini DR and OpenAI DR, while the open-source baselines consist of two multi-agent research frameworks: Open Deep Research (ODR)~\citep{LangChainAI_OpenDeepResearch_2025} and DeerFlow~\citep{ByteDance_DeerFlow}.

For fair comparison, all methods are evaluated under identical tool constraints. In the \textit{Public Search Tool} setting, all methods are restricted to web search only. In the \textit{Public + Internal Search Tools} setting, we extend ODR and DeerFlow with the same internal tool interfaces used by our system. This ensures that performance differences primarily reflect architectural choices rather than disparities in tool availability.

For evaluation on DeepResearch Bench, we run our system in a simplified setting: a single public researcher agent equipped with a web-search tool. We compare against the same LLM-based baselines described above, as well as the top three methods on the benchmark leaderboard—Salesforce AIR~\citep{prabhakar2025enterprisedeepresearchsteerable}, Tavily Research~\citep{GriffSacoranskyNefsky2025BuildingDeepResearch}, and ThinkDepth.ai~\citep{li2025websailornavigatingsuperhumanreasoning}.

In result tables, the notation \textbf{Our}$_{x}$ denotes our DR system instantiated with a specific backbone LLM. Specifically, \textbf{Our}$_{4.1m}$ uses \texttt{gpt-4.1-mini}, \textbf{Our}$_{4.1}$ uses \texttt{gpt-4.1}, and \textbf{Our}$_{5.1}$ uses \texttt{gpt-5.1}.

\subsection{Evaluation Protocol}
\paragraph{Sales Enablement:} We follow a structured evaluation framework, similar to~\citet{du2025deepresearch,venkit2025deeptrace,narayanan2025search,wang2025liveresearchbenchlivebenchmarkusercentric}, that measures DR report quality using four metrics: \textbf{Coverage} \textit{(cov.)}, \textbf{High Actionability Answers} \textit{(HAA)}, \textbf{Readability} \textit{(read.)}, and \textbf{Internal Information Richness} \textit{(IIR)}. Coverage assesses, using a checklist of \textbf{124 evaluation questions}, how comprehensively a report addresses task-relevant information needs. HAA measures the insight depth of covered questions by capturing the proportion of answers that reach a high level of actionability and practical usefulness. Readability evaluates structural clarity, coherence, and linguistic quality of the report. IIR measures how effectively the report incorporates internal knowledge to contextualize externally retrieved information.
%
Full definitions, annotation procedures, and scoring details are in Appendix~\ref{sec:appendix-eval}.

\begin{table}[t]
\centering
\small
\begin{tabular}{lcccc}
\toprule
\textbf{Model} & \textbf{HAA} & \textbf{Cov.} & \textbf{Read.} & \textbf{IIR} \\
\midrule
\multicolumn{5}{c}{{LLM with Public Search Tool}} \\
\midrule
GPT-4.1 & 3.4 & 1.44 & 3.92 & 0.08 \\
GPT-5.1 & 51.90 & 3.13 & 4.06 & 0.13 \\
\midrule
\multicolumn{5}{c}{{Deep Research Systems}} \\
\midrule
\multicolumn{5}{c}{\textit{Public Search Tool}} \\
\midrule
Gemini & 60.00 & 3.51 & \textbf{3.94} & 0.58 \\
OpenAI & 59.67 & 3.53 & 3.84 & 0.54 \\
ODR $_{4.1}$ & 46.36 & 3.18 & 3.70 & 0.44 \\
DeerFlow $_{4.1}$ & 49.80 & 3.30 & 3.70 & 0.48 \\
Our $_{4.1}$ & 71.94 & 3.99 & 3.69 & 0.55 \\ 
Our $_{5.1}$ & 64.60 & 3.70 & 3.90 & 0.56 \\ \midrule
\multicolumn{5}{c}{\textit{Public + Internal Search Tools}} \\
\midrule
ODR $_{4.1}$ & 54.82 & 3.41 & 3.71 & 0.64 \\
DeerFlow $_{4.1}$ & 57.60 & 3.54 & 3.70 & 0.70 \\
Our $_{4.1m}$ & 73.22 & 4.01 & 3.67 & 0.84 \\
Our $_{4.1}$ & \textbf{82.09} & \textbf{4.31} & 3.70 & \textbf{0.89} \\
Our $_{5.1}$ & 72.58 & 4.01 & 3.90 & 0.87 \\
\bottomrule
\end{tabular}
\caption{Results on the enterprise sales enablement task.}
\label{tab:sales-main-results}
\end{table}

\begin{table*}[t]
\centering
\small
\begin{tabular}{lccccc}
\toprule
\textbf{Model} & \textbf{Avg.} & \textbf{Comp.} & \textbf{Ins.} & \textbf{IF} & \textbf{Read.} \\
\midrule
\multicolumn{6}{c}{{LLM with Public Search Tool}} \\
\midrule
GPT-4.1 & 29.31 & 25.59 & 18.42 & 40.63 & 36.49 \\
GPT-5.1 & 51.51 & 50.72 & 51.32 & 52.27 & 51.76 \\
\midrule
\multicolumn{6}{c}{{Deep Research Systems}} \\
\midrule
Salesforce AIR & 50.65 & 50.00 & 51.09 & 50.77 & \textbf{50.32} \\
ThinkDepth.ai & 52.43 & 52.02 & 53.88 & \textbf{52.04} & 50.12 \\
Tavily Research & 52.44 & 52.84 & 53.59 & 51.92 & 49.21 \\
\midrule
Our $_{4.1m}$ & 43.46 & 44.52 & 40.35 & 46.69 & 42.99 \\
Our $_{4.1}$ & 47.02 & 47.95 & 45.09 & 48.81 & 46.72 \\
Our $_{5.1}$ & \textbf{53.40} & \textbf{54.85} & \textbf{55.63} & 50.82 & 49.84 \\
\bottomrule
\end{tabular}
\caption{Results on the DeepResearch benchmark.}
\label{tab:deep-research-bench}
\end{table*}

\paragraph{DeepResearch Bench:} We adopt the RACE protocol and use Comprehensiveness \textit{(comp.)}, Insight/Depth \textit{(ins.)}, Instruction Following \textit{(IF)}, and Readability \textit{(read.)} metrics~\citep{du2025deepresearch}. We use Gemini-2.5-Pro as the LLM judge, following prior findings demonstrating strong agreement between Gemini-2.5-Pro judgments and expert human evaluation~\citep{du2025deepresearch}.


\subsection{Results}

Tables~\ref{tab:sales-main-results} and~\ref{tab:deep-research-bench} present results on the enterprise sales enablement task and the public DeepResearch benchmark, respectively.

\textbf{Our DR system consistently outperforms the underlying LLM backbones.} Compared to LLM-only baselines (GPT-4.1 and GPT-5.1 with search), our system achieves higher scores across both tasks, with particularly strong gains in \emph{comprehensiveness} and \emph{insightfulness}. These improvements are more pronounced for the weaker backbone (GPT-4.1), suggesting that our framework is especially effective when the base model is limited. Although gains over the stronger backbone (GPT-5.1) are smaller, our method still delivers consistent improvements, indicating that structured orchestration and tool integration benefit even more capable models.

\textbf{Our DR system achieves state-of-the-art performance among DR baselines.} On the \emph{enterprise sales enablement} task using \emph{public search tools}, Our $_{4.1}$ achieves the best \emph{coverage} (3.99) and the highest \emph{HAA} (71.94), reflecting stronger evidence gathering and synthesis than competing DR systems. On \emph{DeepResearch Bench}, our system also outperforms leading external methods, including Salesforce AIR, ThinkDepth.ai, and Tavily Research, ranking first in both \emph{comprehensiveness} and \emph{insightfulness}. In particular, Our $_{5.1}$ attains the top \emph{comprehensiveness} (54.85) and \emph{insightfulness} (55.63) scores, consistent with more thorough reasoning over collected information.

\textbf{Internal tools further widen our lead on the sales enablement task.} In the public+internal setting, Our $_{4.1}$ achieves the strongest overall results, with the highest \emph{coverage} (4.31) and \emph{HAA} (82.09), improving over the public-only setting (3.99 coverage, 71.94 HAA). It also leads on \emph{IIR}, which measures how effectively internal search evidence is incorporated into the final report: Our $_{4.1}$ scores 0.89 versus 0.70 for DeerFlow and 0.64 for ODR, indicating more effective integration of internal knowledge with external sources.

\textbf{Templates improve enterprise DR by constraining decisions and reducing dependence on the strongest reasoning models.}
On sales enablement, Our $_{4.1}$ performs best, with Our $_{5.1}$ and Our $_{4.1m}$ close behind. On DeepResearch Bench, performance more clearly tracks model capacity: Our $_{5.1}$ outperforms Our $_{4.1}$, which outperforms Our $_{4.1m}$. We attribute this difference to how tightly each task constrains the research process. DeepResearch Bench provides no fixed structure and a less-specified target response, so the system must decide what to cover and how to organize it, rewarding stronger reasoning. Sales enablement, by contrast, uses a fixed report template that pre-defines key sections and reduces planning ambiguity. In this setting, the more focused Our $_{4.1}$ produces more aligned reports, whereas Our $_{5.1}$ sometimes expands beyond the template and adds context that is not required.


\begin{table*}[t]
\centering
\small
\begin{tabular}{lccccccc}
\toprule
\textbf{Model} & \textbf{HAA} & \textbf{Cov.} & \textbf{Read.} & \textbf{IIR} & \textbf{Time (min)} & \textbf{\#Tool Public} & \textbf{\#Tool Internal} \\
\midrule
Our $_{4.1}$ & 82.09 & 4.31 & 3.70 & 0.89 & 47 & 327 & 90 \\
Sequential (no plan DAG) & 76.40 & 4.10 & 3.70 & 0.85 & 222 & 368 & 112 \\
\midrule
\quad w/o Agent Termination & 73.67 & 4.02 & 3.70 & 0.84 & 43 & 224 & 64 \\
\quad w/o Outline Reflection & 70.75 & 3.94 & 3.65 & 0.84 & 42 & 216 & 65 \\
\quad w/o Outline & 67.30 & 3.80 & 3.64 & 0.82 & 40 & 198 & 57 \\
\bottomrule
\end{tabular}
\caption{Ablation study examining the role of dependency-aware planning and explicit completion criteria in report quality, coverage, and efficiency for the sales enablement task.}
\label{tab:ablation-sales-results}
\end{table*}

\subsection{Ablation Study} \label{sec:ablation}
\Cref{tab:ablation-sales-results} presents an ablation study on the sales enablement task, analyzing the contribution of key system components to both performance and efficiency.

\textbf{Dependency-aware planning and selective context sharing improve both report quality and research efficiency.} 
We first evaluate a sequential variant that generates a plan but omits an explicit dependency graph. Steps are executed strictly one at a time, and the context for each step is the cumulative output of all prior steps. Removing the Plan DAG substantially degrades both quality and efficiency: HAA drops from 82.09 to 76.40 and coverage falls from 4.31 to 4.10, while execution time increases from 47 to 222 minutes. The slowdown coincides with heavier tool use: public search calls rise from 327 to 368 and internal search calls from 90 to 112, consistent with the loss of dependency-aware parallelism and reduced reuse of intermediate results.

\textbf{Removing explicit completion criteria causes large drops in quality and coverage.}
We next ablate components that govern ``done-ness.'' Removing agent termination yields the largest degradation: HAA falls from 82.09 to 73.67 and coverage from 4.31 to 4.02. This variant also stops earlier, with tool calls dropping from 327 to 224 (public) and from 90 to 64 (internal), indicating premature termination before sufficient evidence is gathered. Removing outline reflection and outline generation further reduces performance: coverage declines to 3.94 and 3.80, and HAA to 70.75 and 67.30, respectively. Overall, these components operationalize what ``done'' means: completion criteria at both the plan level and the step level are essential to prevent premature stopping and to produce high-coverage, actionable reports.

\section{Conclusion}
We presented an Enterprise Deep Research system that improves reliability by making information needs, information flow, and completion decisions explicit. The system (i) decomposes requests into coverage-driven objectives with reflective auditing, (ii) executes them via a dependency-structured plan with localized contexts and dependency-gated sharing, and (iii) applies evidence-based sufficiency criteria to reduce premature stopping and stabilize depth across sections. Experiments on internal sales enablement win-card generation and the public DeepResearch Bench show improved coverage and report quality versus competitive baselines, with ablations indicating that dependency-controlled context sharing and step-level sufficiency criteria drive most of the gains.

\section*{Limitations}
For win-card evaluation, we partnered with real account executives, the intended end users of these reports, to ensure the assessment reflects practical sales needs. The findings are specific to the win-card setting, since building reliable evaluation setups for other enterprise tasks requires substantial domain effort. At the same time, our strong performance on the public DeepResearch Bench benchmark suggests the core design generalizes beyond a single internal task. We recommend applying the approach to new enterprise tasks alongside a tailored evaluation protocol to verify quality, coverage, and decision-readiness in the target use case.



\bibliography{custom}

\appendix

\section{Evaluation Framework for Sales Enablement Task}\label{sec:appendix-eval}

Following recent work \cite{du2025deepresearch, venkit2025deeptrace, narayanan2025search, wang2025liveresearchbenchlivebenchmarkusercentric} on deep research agents, we adopt a structured evaluation framework that decomposes output quality into three primary dimensions: Coverage, Readability, and Internal Information Richness (IIR). Within Coverage, we additionally report High Actionability Answers (HAA) as a subcomponent that captures the depth and practical usability of insights within covered content. Each dimension is defined through an explicit protocol and evaluated using a combination of human annotation and LLM-based judges.

\subsection{Coverage}

Coverage measures \textit{whether a generated research report addresses the full set of information needs relevant to the sales enablement task context}. To evaluate coverage, we first constructed an initial set of 90 coverage questions based on domain expertise and external company-specific data, covering strategic, business, and technical aspects that a sales enablement report should address. We then expanded this pool with an additional 83 questions generated using LLM-based analysis of deep research reports produced by OpenAI, Gemini, and DeerFlow systems. This resulted in a combined set of 173 candidate questions.

To refine this set, we conducted a relevance annotation study with three Account Executives (AEs) with at least two years of experience in technology and CRM systems, recruited via UserInterviews. Annotators independently labeled each question as either relevant or non-relevant and suggested missing questions. Only questions with unanimous agreement on relevance were retained, resulting in a set of 104 questions. Additional AE-suggested questions were added after removing duplicates, yielding a final set of 124 questions.

These questions serve as a checklist for evaluation. Coverage is measured by scoring how well each question is addressed in the generated report using a five-point Likert scale that reflects the depth and usefulness of the answer for sales enablement.

A score of \textbf{1 (No Answer)} indicates that the report provides no relevant information, is too generic, outdated, or off-topic, and offers no actionable detail for a sales conversation. A score of \textbf{2 (Weak / Generic)} indicates that the report mentions broad themes or priorities but lacks customer-specific details, ownership, or timeframe, requiring substantial additional research by the AE. A score of \textbf{3 (Partial / Usable)} indicates that the question is addressed with some customer-specific information, but key details such as timelines, sponsors, KPIs, or specificity are missing, meaning the AE would need to supplement the information before a meeting. A score of \textbf{4 (Strong but Incomplete)} indicates that most aspects are covered with specific details such as initiatives, timelines, or metrics, but one or two important elements are missing. A score of \textbf{5 (Fully Covered)} indicates complete, actionable, and customer-specific coverage, including relevant names, numbers, timelines, KPIs, and context, making the information ready to use directly in a customer meeting.

We report two coverage-derived measures. The first is the \textbf{average coverage score}, computed as the mean score across all evaluation questions and reflecting the breadth of information addressed by the report. The second is \textbf{High Actionability Answers (HAA)}, a subcomponent of coverage defined as the percentage of questions receiving a score of 4 or higher. HAA captures the proportion of answers that are sufficiently detailed and actionable for practical use, reflecting the depth and operational usefulness of covered content.

\subsection{Readability}

Readability evaluates \textit{whether a generated report is easy to follow, well structured, and appropriate for its intended audience}. We decompose readability into five components commonly used in discourse-level evaluation \cite{du2025deepresearch}: structure and information flow, language fluency and style, lexical and syntactic complexity, cohesion and coherence, and presentation and formatting. Each component is scored on a five-point Likert scale, and the final readability score is computed as their unweighted average.

\subsection{Internal Information Richness}

Internal Information Richness (IIR) measures \textbf{how effectively the report incorporates internal Salesforce knowledge to enrich and contextualize externally retrieved information}, rather than presenting external facts in isolation. IIR rewards reports that use internal context to interpret ambiguous signals, connect findings to account-specific realities, and produce actionable, customer-tailored analysis.

IIR is first scored on a 0-100 scale using three weighted components:

\textbf{1. Internal Access Advantage (0-50 points).}  
This component measures the presence of signals that indicate access to internal Salesforce knowledge. Points are awarded as follows:
\begin{itemize}[leftmargin=*]
\item Deal signals: 5 points each, up to 10 points
\item Named champions: 4 points each, up to 12 points
\item Internal tool traces: 4 points each, up to 8 points
\item Internal architecture details: 3 points each, up to 6 points
\item Confirmed product usage: 2 points each, up to 6 points
\item Internal strategic context: 2 points each, up to 4 points
\item Internal terminology usage: 1 point each, up to 4 points
\end{itemize}

The total for this section is capped at 50 points.

\textbf{2. Internal Knowledge Depth (0-30 points).}  
This component evaluates whether the report demonstrates deep, multi-layered understanding of the customer environment. Points are awarded as follows:
\begin{itemize}[leftmargin=*]
\item Multi-org complexity: 6 points (max 1 instance)
\item Implementation roadmap detail: 6 points (max 1 instance)
\item Product PoC details: 5 points each, up to 10 points
\item Strategic classification (e.g., replacement vs. greenfield): 4 points (max 1 instance)
\item Competitive intelligence: 2 points (max 1 instance)
\item Cross-cloud integration detail: 2 points (max 1 instance)
\end{itemize}  

The total for this section is capped at 30 points.

\textbf{3. Ground-Truth Alignment (0-20 points).}  
This component measures alignment with internally verifiable account realities. It consists of five binary checks, each worth 4 points:
\begin{itemize}[leftmargin=*]
\item Accurate identification of owned Salesforce products and licenses
\item Recognition of multi-org complexity as a core challenge
\item Confirmation of competitor contract status using internal knowledge
\item Identification of named internal champions or prior stakeholders
\item Evidence of past Salesforce interactions (e.g., trials or PoCs)
\end{itemize}

Each satisfied condition earns 4 points, for a maximum of 20 points.

\textbf{Final Score.}  
The IIR score is the sum of all three components:
\begin{itemize}[leftmargin=*]
\item Internal Access Advantage (max 50)
\item Internal Knowledge Depth (max 30)
\item Ground-Truth Alignment (max 20)
\end{itemize}

This yields a raw score out of 100. The final IIR score is then normalized to the range \([0,1]\) by dividing the raw score by 100, reflecting how effectively internal knowledge is used to enhance interpretation, accuracy, and strategic relevance. 

\section{Examples}
\subsection{Outline} \label{appendix-outline}
\begin{tcolorbox}[
  title={Task \& Outline},
  colback=white,
  colframe=black!60,
  fonttitle=\bfseries,
  breakable
]
\small
\textbf{Task:}\\
From 2020 to 2050, how many elderly people will there be in Japan? What is their consumption potential across various aspects such as clothing, food, housing, and transportation? Based on population projections, elderly consumer willingness, and potential changes in their consumption habits, please produce a market size analysis report for the elderly demographic.

\vspace{0.8em}
\textbf{Outline:}

\vspace{0.4em}
\textbf{I. What are the projected demographic and economic trends of Japan’s elderly population (aged 65+) from 2020 to 2050, and how do these interact to shape consumption potential?}

\begin{enumerate}[label=\Alph*.]
  \item What are the overall population projections for Japan's elderly by decade (2020, 2030, 2040, 2050)?
  \begin{enumerate}[label=\arabic*.]
    \item What is the expected absolute number of elderly (65+) in each reference year?
    \item What is the projected percentage of elderly relative to the total Japanese population in each decade?
    \item What is the annualized growth or decline rate of the elderly demographic across each decade?
    \item How have elderly population trends evolved pre-2020, and what is the trajectory for 2020--2050?
  \end{enumerate}

  \item How are demographic variables expected to change within the elderly group?
  \begin{enumerate}[label=\arabic*.]
    \item How will gender distribution (male vs.\ female) within the elderly population shift per decade?
    \item What are the projected geographical distributions (urban vs.\ rural; by prefecture or region) of the elderly, and how will regional trends diverge or converge over time?
    \item How will age subgroups evolve within the elderly (e.g., 65--74, 75--84, 85+) over time?
    \item What is the expected life expectancy for the elderly by gender in each decade?
  \end{enumerate}

  \item How will trends in household composition and living arrangements affect the elderly market?
  \begin{enumerate}[label=\arabic*.]
    \item What proportions of elderly are expected to live alone, with spouse/family, or in institutions (e.g., senior living facilities, nursing homes) in each decade?
    \item How do these living arrangements differ by age, gender, region, or income level?
    \item How will changes in household composition influence sectoral consumption potential (clothing, food, housing, transportation)?
  \end{enumerate}

  \item What are the primary factors influencing elderly demographic and economic projections?
  \begin{enumerate}[label=\arabic*.]
    \item What are projected trends for fertility, mortality, and migration affecting the elderly segment?
    \item How might government policies, including retirement age, healthcare systems, and immigration, influence the elderly demographic size and structure?
    \item How will projected changes in pension systems, social security reforms, or eligibility ages affect elderly disposable income and overall market size?
    \item How will shifts in elderly labor force participation or prolonged workforce engagement interact with consumption potential?
  \end{enumerate}

  \item How does projected elderly growth compare to other age groups and past trends?
  \begin{enumerate}[label=\arabic*.]
    \item How does projected elderly consumption and population size compare with the working-age (15--64) and youth (0--14) populations in each decade?
    \item What are the implications for overall market structure and sectoral shifts?
  \end{enumerate}
\end{enumerate}

\vspace{0.8em}
\textbf{II. What is the estimated market size and consumption potential of Japan’s elderly (65+) in key sectors from 2020 to 2050?}

\begin{enumerate}[label=\Alph*.]
  \item What are the estimated market sizes (in JPY or USD) for each sector (clothing, food, housing, transportation) in the base year (2020)?
  \begin{enumerate}[label=\arabic*.]
    \item What are the data sources and methodologies (e.g., per capita spending, consumption surveys) used for establishing the 2020 baseline?
    \item What are the estimated sector market sizes by demographic subgroup (age band, gender, income, region) in 2020?
  \end{enumerate}

  \item What is the methodology for projecting sector market sizes from 2020 to 2050?
  \begin{enumerate}[label=\arabic*.]
    \item What estimation models, key assumptions (e.g., per capita consumption growth, inflation or deflation per sector), and demographic drivers are used for forecasting?
    \item How are sector-specific price trends and inflation/deflation incorporated into market size projections?
    \item What are the scenarios or sensitivity analyses for key variables (e.g., economic growth, policy changes, shocks)?
  \end{enumerate}

  \item What are the projected market sizes for the elderly in each sector and sub-sector for each decade?
  \begin{enumerate}[label=\arabic*.]
    \item \textbf{Clothing:} What is the total and subgroup consumer spending on clothing, and how is demand expected to be distributed across types or brands?
    \item \textbf{Food:} How is spending distributed among groceries, prepared/ready-to-eat meals, and dining out, and how do shares shift over time?
    \item \textbf{Housing:} How will demand be split among home ownership, rental apartments, senior living facilities, and institutional care; what are the value and unit demand in each?
    \item \textbf{Transportation:} How do public transport, private vehicles, and emerging mobility services (e.g., ride-sharing, autonomous shuttles) factor into and change the overall market size?
    \item For each sector, what is the breakdown by age subgroup (65--74, 75--84, 85+), gender, region, income, household type, and living arrangement?
  \end{enumerate}

  \item To what extent do market sizes in each sector reflect regional disparities and urban/rural divides?
  \begin{enumerate}[label=\arabic*.]
    \item What are the current and projected regional (e.g., by prefecture, metropolitan area) market sizes in each sector?
    \item How do regional time trends---such as regional divergence or convergence---shape overall market opportunities?
    \item How does access to sector-relevant goods and services (e.g., proximity to supermarkets or transport networks) affect regional sector potential?
  \end{enumerate}

  \item How do market entry barriers or facilitators influence sector growth for elderly consumers?
  \begin{enumerate}[label=\arabic*.]
    \item What regulatory, infrastructural, or supply-side factors (e.g., housing policy, transport regulation, healthcare licensing) might constrain or promote elderly market expansion in each sector?
    \item How might ecosystem development or new business models unlock untapped consumption potential?
  \end{enumerate}

  \item How does the penetration of new goods/services and digital adoption affect future market sizes?
  \begin{enumerate}[label=\arabic*.]
    \item What are the projected adoption and penetration rates of technology-enabled goods/services (e.g., smart homes, online food/grocery shopping, telemedicine, mobility-as-a-service) among the elderly, especially in 2040--2050?
    \item What are the barriers and catalysts for digital and technological adoption in each sector?
  \end{enumerate}

  \item How does sector-specific inflation/deflation impact real consumption capacity and market growth?
  \begin{enumerate}[label=\arabic*.]
    \item What are the trends in price indices for clothing, food, housing, and transportation?
    \item How does cost growth, especially in essentials (e.g., housing, healthcare, food), affect elderly real purchasing power?
    \item What is the likely influence of public subsidies, sector regulations, or consumer protection on affordability?
  \end{enumerate}
\end{enumerate}

\vspace{0.8em}
\textbf{III. How do changes in elderly consumer behavior, willingness, and external factors influence market size projections over time?}

\begin{enumerate}[label=\Alph*.]
  \item How does elderly consumer willingness to spend evolve from 2020 to 2050?
  \begin{enumerate}[label=\arabic*.]
    \item How do changes in disposable income, pension coverage, and wealth impact willingness and ability to spend in each sector?
    \item What are the projected elderly saving versus spending rates for each decade?
    \item How do psychological factors (e.g., perceptions of health, longevity, economic security) influence consumption behavior by sector?
  \end{enumerate}

  \item How are elderly consumption habits, needs, and sector preferences evolving?
  \begin{enumerate}[label=\arabic*.]
    \item How are generational effects (e.g., wartime vs.\ baby boomer cohorts) and societal shifts altering consumption patterns, preferences for brands, convenience, or sustainability?
    \item How do health trends, mobility limitations, or the need for care services alter consumption of clothing, food, housing, and transportation?
    \item How will the growth of single-person elderly households and institutionalization influence goods/services demand?
    \item What is the projected uptake of novel products/services (e.g., smart appliances, AI-enabled eldercare, mobility aids, prepared meal delivery) by different elderly segments and over time?
  \end{enumerate}

  \item How do external macroeconomic and policy factors affect elderly market size projections?
  \begin{enumerate}[label=\arabic*.]
    \item What are the effects of macroeconomic shifts---recessions, inflation, pension system shocks---on sectoral consumption among the elderly?
    \item How might government or private sector policies (retirement age, eldercare support, subsidies, tax code revisions) alter affordability and sectoral spending patterns?
    \item What is the potential impact of significant social and technological events (e.g., pandemics, climate disasters, breakthroughs in assistive technology) on elderly consumption, both positive and negative?
  \end{enumerate}

  \item How do cross-sector and cross-demographic interactions shape the future elderly market?
  \begin{enumerate}[label=\arabic*.]
    \item To what extent do shifts in elderly consumption displace or complement that of younger age groups, and how might intergenerational transfers (inheritance, family support) influence market dynamics?
    \item How could the evolution of elderly consumption patterns catalyze new market opportunities, supply chain adaptations, or disruptive innovation?
  \end{enumerate}
\end{enumerate}

\end{tcolorbox}

\subsection{Plan DAG} \label{appendix-plan-dag}

\begin{tcolorbox}[
  title={Task \& Plan DAG},
  colback=white,
  colframe=black!60,
  fonttitle=\bfseries,
  breakable
]

\textbf{Task:}\\
Research on the price dynamics of chub mackerel in major aquatic markets of Pacific Rim countries, and its interannual variations in weight/length. Combined with oceanographic theory, these research findings can further establish direct correlations between high-quality marine biological resources, aquatic markets, fishery economics, and the marine environment.

\vspace{0.8em}
\textbf{Plan DAG:}

\begin{itemize}[leftmargin=*, itemsep=0.5em]
  \item \textbf{Locale:} en-US
  \item \textbf{Task statement:} Investigate the price dynamics of chub mackerel in major aquatic markets of Pacific Rim countries, analyze interannual variations in weight/length, and integrate oceanographic theory to examine correlations between marine biological resources, aquatic markets, fishery economics, and the marine environment.
  \item \textbf{Completion flag:} task\_completed = false
\end{itemize}

\vspace{0.4em}
\textbf{Step Groups:}

\begin{enumerate}[leftmargin=*, label=\arabic*. , itemsep=0.8em]

  \item \textbf{Map Pacific Rim chub mackerel markets and price/value chain dynamics across countries and over time.}
  \begin{itemize}[leftmargin=*, itemsep=0.35em]
    \item \textbf{S01-01 (research, pending; deps: ---):} Identify primary Pacific Rim countries and major aquatic markets involved in chub mackerel trade; determine market centers, volume leaders, and export/import nodes using FAO, UN Comtrade, government fisheries reports, and market authority sources.
    \item \textbf{S01-02 (research, pending; deps: S01-01):} Collect landing/wholesale/retail price ranges (avg/min/max) across identified markets for the last 10--15 years; capture domestic vs.\ export shares where possible.
    \item \textbf{S01-03 (research, pending; deps: S01-01, S01-02):} Analyze spatial/temporal price trends and shocks; assess geographic differentials and responses to supply shocks (seasonality, ocean events, regulation) using time-series analysis and fisheries news archives.
    \item \textbf{S01-04 (research, pending; deps: S01-01):} Research value-chain structure (actors, transaction systems, grading/quality standards, logistics, trade rules/tariffs), distinguishing domestic vs.\ export flows.
    \item \textbf{S01-05 (research, pending; deps: S01-01, S01-02):} Compare chub mackerel prices with substitute species (e.g., horse mackerel, sardine, other small pelagics); assess co-movement using price series and industry/nutrition market reports.
  \end{itemize}

  \item \textbf{Assess interannual and spatial variations in chub mackerel biological metrics (weight, length, quality) and their market relevance.}
  \begin{itemize}[leftmargin=*, itemsep=0.35em]
    \item \textbf{S02-01 (research, pending; deps: ---):} Collect datasets on weight/length/age/condition from fisheries agencies, surveys, and literature across Pacific Rim fishing grounds; include zones, seasons, inshore/offshore differences.
    \item \textbf{S02-02 (research, pending; deps: S02-01):} Assess comparability and gaps in biological data across countries/years; identify harmonization efforts and key inconsistencies (international survey methods, FAO technical papers).
    \item \textbf{S02-03 (research, pending; deps: S02-01, S02-02):} Analyze interannual/regional changes in size metrics; link market grading to survey size/condition; use LMEM/von Bertalanffy models where available.
    \item \textbf{S02-04 (research, pending; deps: S02-01):} Investigate drivers of variation (age structure, fishing intensity, recruitment, management regimes, definitions of quality) using stock assessments and market grading standards.
  \end{itemize}

  \item \textbf{Analyze oceanographic and environmental influences on chub mackerel biology and markets, and their quantitative integration.}
  \begin{itemize}[leftmargin=*, itemsep=0.35em]
    \item \textbf{S03-01 (research, pending; deps: ---):} Compile key oceanographic variables (SST, upwelling, productivity, oxygenation, marine heatwaves, ENSO, hypoxia, pollutants) using NOAA/JAMSTEC/IMOS/Copernicus and peer-reviewed studies; summarize magnitude/variation/impacts.
    \item \textbf{S03-02 (research, pending; deps: ---):} Review modeling frameworks integrating biological, oceanographic, and market data (multivariate, regression, time-series; GTWR, LMEM, bioeconomic models, SEM); evaluate approaches for causality/feedback.
    \item \textbf{S03-03 (research, pending; deps: S03-01, S03-02):} Investigate quantitative links between ocean variables and biology/market outcomes; identify mechanisms and statistical associations tied to shifts in quality and prices.
    \item \textbf{S03-04 (research, pending; deps: S01-03, S02-03, S03-02):} Identify how confounders (policy/regulation, anomalies, non-environmental shocks) are controlled for in integrative analyses using methodological reviews and controlled model studies.
  \end{itemize}

  \item \textbf{Contextualize biological, market, and environmental results: socioeconomic, policy, and regional mediation.}
  \begin{itemize}[leftmargin=*, itemsep=0.35em]
    \item \textbf{S04-01 (research, pending; deps: ---):} Profile demographic/economic/regulatory characteristics shaping exploitation and trade (employment, food security, export earnings, resilience/vulnerability) using national reports, World Bank/FishStat, and development studies.
    \item \textbf{S04-02 (research, pending; deps: ---):} Review fisheries management systems (TAC/quota, ITQs, closures, size limits, certifications, enforcement); evaluate impacts on biological and market outcomes.
    \item \textbf{S04-03 (research, pending; deps: ---):} Examine value-chain orientation (local vs.\ export) and market/process innovations (processing, freezing, auctions) affecting prices and competitiveness; synthesize sector analyses.
  \end{itemize}

  \item \textbf{Synthesize applied management implications and future research needs for integrating biological, market, and environmental knowledge.}
  \begin{itemize}[leftmargin=*, itemsep=0.35em]
    \item \textbf{S05-01 (research, pending; deps: S01-03, S02-03, S03-03, S04-02):} Identify adaptive management strategies, policy innovations, and sustainability interventions emerging from integrated analysis (best practices, incentives, certifications, resilience/vulnerability studies).
    \item \textbf{S05-02 (research, pending; deps: S01-03, S02-03, S03-03, S04-03):} Assess data/knowledge gaps and needs for monitoring/modeling/harmonization; compile recommendations for interdisciplinary and international collaboration from white papers and summit outcomes.
  \end{itemize}

\end{enumerate}

\end{tcolorbox}
\twocolumn

\subsection{Agent Termination Criteria} \label{appendix-termination-criteria}

\begin{tcolorbox}[
  title={Step \& Termination Criteria},
  colback=white,
  colframe=black!60,
  fonttitle=\bfseries,
  breakable
]
\small
\textbf{Step:}\\
Collect official population projections for Japan’s elderly (aged 65+) for the years 2020, 2030, 2040, and 2050 from sources such as the National Institute of Population and Social Security Research, Japan Statistics Bureau, and United Nations Population Division. Record absolute numbers, percentages of total population, decade growth/decline rates, and historical context (pre-2020 trends and post-2020 trajectory).

\vspace{0.8em}
\textbf{Termination Criteria:}

\begin{itemize}[leftmargin=*, itemsep=0.35em]
  \item \textbf{Coverage specification:} Official population projections for Japan’s elderly (aged 65+) must be provided for the years \textbf{2020, 2030, 2040, and 2050}.
  \item \textbf{Each data point must include:}
  \begin{itemize}[leftmargin=*, itemsep=0.25em]
    \item Absolute number of elderly (aged 65+) for each year.
    \item Percentage of elderly (aged 65+) as a share of total population for each year.
    \item Growth or decline rate of elderly (aged 65+) between each decade (2020--2030, 2030--2040, 2040--2050), expressed as a percentage.
  \end{itemize}

  \item \textbf{Sources must include, at minimum, published data or projections from:}
  \begin{itemize}[leftmargin=*, itemsep=0.25em]
    \item National Institute of Population and Social Security Research (Japan)
    \item Japan Statistics Bureau
    \item United Nations Population Division
  \end{itemize}

  \item Discrepancies or variances between sources must be explicitly noted.

  \item \textbf{Provide a summary of historical context:}
  \begin{itemize}[leftmargin=*, itemsep=0.25em]
    \item High-level narrative of pre-2020 trends in Japan’s elderly population (e.g., demographic shifts, major changes in growth rates), with quantitative references where possible.
    \item Outline and reference the projected post-2020 trajectory to 2050, highlighting key inflection points or notable patterns from the projections.
  \end{itemize}

  \item All figures must be precisely attributed to the originating source and use the latest available official projection sets as of the research cutoff.

  \item Coverage is national only; no subnational (e.g., prefectural, city) or international breakdowns are required or permitted.

  \item Only projections/models produced or published by the three specified organizations are valid; exclude unofficial estimates, academic papers, news reports, or secondary data compilers.

  \item The elderly population must be defined as aged 65 and over; any other definitions or segmentations (e.g., 75+, 85+) are excluded unless directly relevant to calculating totals for 65+.
\end{itemize}

\end{tcolorbox}
\twocolumn




\end{document}